\documentclass{article}

\usepackage[preprint]{template}

\usepackage[utf8]{inputenc} 
\usepackage[T1]{fontenc}    
\usepackage{hyperref}       
\usepackage{url}            
\usepackage{booktabs}       
\usepackage{amsfonts}       
\usepackage{nicefrac}       
\usepackage{microtype}      
\usepackage{xcolor}         

\usepackage{amsmath}

\usepackage[misc]{ifsym}
\usepackage{graphicx}
\usepackage{multirow}

\title{Evaluating Post-hoc Explanations of the Transformer-based Genome Language Model DNABERT-2}

\author{%
    Isabel Kurth\thanks{I. Kurth and P. Yanez Sarmiento --- Equal contribution.}\\
    Hasso Plattner Institute\\
    University of Potsdam\\
    \texttt{isabel.kurth@student.hpi.de}\\
    \And
  Paulo~Yanez Sarmiento$^*$ \\
  Hasso Plattner Institute\\
  University of Potsdam\\
  \texttt{paulo.yanez@hpi.de}\\
  \AND
  Bernhard Y. Renard (\Letter)\\
  Hasso Plattner Institute\\
  University of Potsdam\\
  \texttt{bernhard.renard@hpi.de}\\
}

\begin{document}

\maketitle

\begin{abstract}
Explaining deep neural network predictions on genome sequences enables biological insight and hypothesis generation---often of greater interest than predictive performance alone. While explanations of convolutional neural networks (CNNs) have been shown to capture relevant patterns in genome sequences, it is unclear whether this transfers to more expressive Transformer-based genome language models (gLMs). To answer this question, we adapt AttnLRP, an extension of layer-wise relevance propagation to the attention mechanism, and apply it to the state-of-the-art gLM DNABERT-2. Thereby, we propose strategies to transfer explanations from token and nucleotide level. We evaluate the adaption of AttnLRP on genomic datasets using multiple metrics. Further, we provide an extensive comparison between the explanations of DNABERT-2 and a baseline CNN. Our results demonstrate that AttnLRP yields reliable explanations corresponding to known biological patterns. Hence, like CNNs, gLMs can also help derive biological insights. This work contributes to the explainability of gLMs and addresses the comparability of relevance attributions across different architectures.
\\

\end{abstract}

\section{Introduction}\label{sec_introsuction}
In genomics, one common application of deep learning is to classify genome sequences based on regulatory elements such as promoters, enhancers, or transcription factor binding sites \cite{eraslan2019deep,zou2019primer}. In particular, deep learning models are trained on genome sequences that consist of letters of the alphabet $\{\mathrm{A,C,G,T}\}$ where each letter represents one of the four nucleotides present in DNA. For years, convolutional neural networks (CNNs) have been a common deep learning architecture for genomics \cite{eraslan2019deep}. They are capable of learning patterns or motifs from raw sequences and combining them to make prediction. Thereby, not only the predictive performance but also the explainability of the model is of interest for confirming experimental results or hypothesis generation \cite{novakovsky2023obtaining}. In fact, in contrast to other fields such as computer vision, in genomics, leveraging deep learning and explainability for generating new hypotheses is often of higher interest than model inference \cite{bartoszewicz2021interpretable}. For instance, it can help to better understand genome function and regulatory mechanisms and hence, potentially improve diagnosis, treatment, and prevention of diseases \cite{eraslan2019deep,novakovsky2023obtaining}.

There are two types of approaches to gain insights into the inner working of a deep learning model: interpreting the model directly, e.g., analyzing what the filters of a CNN learned, or by applying post-hoc attribution methods (also called explanation methods) \cite{eraslan2019deep,novakovsky2023obtaining}. In this work, we focus on the latter. Given a trained model, post-hoc attribution methods assign relevance scores to every input feature for every sample and thereby provide a local explanation of the model. Layer-wise relevance propagation (LRP) \cite{bach2015pixel,montavon2019layer} is a widely used post-hoc attribution method that propagates relevance backwards from the output of the model to its input. Precisely, it decomposes the model's prediction into relevance scores attributed to every input feature.

In recent years, there has been an emergence of Transformer-based language models trained self-supervised on large datasets of genome sequences: so-called DNA or genome language models (gLM) \cite{consens2025transformers}. The attention layers in Transformer models are capable of capturing long-range dependencies between features and therefore useful for genomics where nucleotides can affect other areas of the DNA that are far away. If they are fine-tuned and equipped with a downstream head, they show promising performance on several downstream tasks. Among them is DNABERT-2 \cite{zhou2023dnabert}, a BERT-stlye gLM trained on a multi-species genome dataset. It outperformed its predecessor DNABERT \cite{ji2021dnabert} and achieves performance comparable to the parameter-wise much larger Nucleotide Transformer \cite{dalla2025nucleotide}.

However, most post-hoc attribution methods---including LRP---are not directly transferable to gLMs and their Transformer architecture due to nonlinearities in the attention mechanism. Recently, AttnLRP has been proposed to explain Transformer-based models and applied to natural language processing and computer vision \cite{achtibat2024attnlrp}. In this work, we adapt and apply AttnLRP to obtain explanations of DNABERT-2 fine-tuned on two classification tasks of a commonly used benchmark dataset \cite{gresova2023genomic}. For the evaluation of the explanations, we consider metrics for sparsity, complexity, similarity, faithfulness, and localization in the form of a database comparison. As a baseline for our analysis, we train a CNN on the same datasets and generate explanations with LRP.

\section{Related Work}\label{sec_related_work}
LRP is a post-hoc attribution method that provides local explanations of a trained deep learning model introduced by \cite{bach2015pixel}. It has been further developed since to address limitations and has been extended to more complex network architectures \cite{montavon2019layer,samek2021explaining}. LRP is related to DeepLIFT which also propagates relevance backwards through the neural network \cite{shrikumar2017learning}. Both belong to a class of explanation methods which can be generalized to SHAP values \cite{lundberg2017unified}. While SHAP values are model agnostic, their exact computation is expensive. Therefore, DeepSHAP extends DeepLIFT to efficiently calculate SHAP values for feed-forward neural networks \cite{lundberg2017unified}. However, it does not naturally transfer to the attention mechanism in Transformer models.

Although there have been approaches before to extend LRP to Transformer models \cite{ali2022xai,chefer2021transformer}, AttnLRP is the first modification that faithfully and efficiently handles the nonlinearities in the attention mechanism \cite{achtibat2024attnlrp}. More recently, \cite{arras2025close} showed that AttnLRP can be computed more efficiently by using a modified Gradient$\times$Input approach compared to the explicit implementation by \cite{achtibat2024attnlrp}.

DNABERT-2 \cite{zhou2023dnabert} and its predecessor DNABERT \cite{ji2021dnabert} are Transformer-based gLMs building upon BERT \cite{devlin2019bert}. DNABERT-2 uses Byte Pair Encoding (BPE) instead of DNABERT's overlapping $k$-mer tokenization to overcome several limitations such as data redundancy and leakage. Further, it implements Attention with Linear Bias (ALiBi) which adds a distance-dependent
bias to avoid positional embedding.

While gLMs achieve state-of-the-art performance on many benchmarks, \cite{marin2024bend} found that, depending on the downstream task, supervised trained CNNs still exhibit competitive performance. \cite{tang2025evaluating} evaluated the transferability of several gLMs representations for different regulatory genomics downstream tasks. They also find that the non-fine-tuned gLM representations did not lead to a substantial performance improvement over a CNN trained on one-hot-encoded sequences. They also considered relevance attributions, derived by masking tokens and considering the change in entropy over the predicted distribution or by saliency maps of a CNN supervised trained on the gLM representations. 
Alternatively, \cite{clauwaert2021explainability} considered the scores of the attention head of a Transformer-based neural network trained on genome sequences directly and could match their focus to known transcription factor binding sites. 

\cite{consens2023transforming} applied a previous modification of LRP by \cite{chefer2021transformer} to the predecessor model DNABERT \cite{ji2021dnabert}. They compare LRP explanations to attention scores and other explanation methods by considering the drop in performance. To the best of our knowledge, AttnLRP has not been applied in the field of genomics and especially, its explanations have not been evaluated and benchmarked against explanations of CNNs.

\section{Method}\label{section_adaption_attnlrp}

We adapt AttnLRP to make it applicable to DNABERT-2. Although 
it works directly with the attention layers of the BERT architecture \cite{devlin2019bert}, DNA\-BERT-2 contains two components, ALiBi and gated linear units (GLU), which require an adaptation of AttnLRP.
For our application of AttnLRP, we follow the modified Gradient$\times$Input approach and the resulting rules proposed by \cite{arras2025close}.
Specifically, for the products in ALiBi and GLU, we also apply the uniform redistribution rule, i.e., the relevance of an output neuron is split equally among its input factors. 
For ALiBi, this implies that the relevance is equally split between query and key separately from the distance-dependent bias. For the GLU, the relevance is equally split between gated and content branches.

AttnLRP assigns relevance to every token which in general consists of multiple nucleotides. In contrast, LRP applied to a CNN provides relevance scores on nucleotide level. Therefore, the explanations of AttnLRP and LRP are not straightforward comparable. Further, DNABERT-2 has two special tokens: the class (CLS) and separation token (SEP). CLS is intended to represent the entire sequence and hence, depends on the other tokens. In contrast, SEP indicates the end of the sequence. In our analysis, while we examine relevance attributed to CLS and SEP, our main interest is the relevance of tokens representing specific parts of the sequence as this allows comparison to the CNN explanations and biological interpretation.

To be able to properly compare and evaluate gLM and CNN explanations, we propose four (dis\nobreakdash-)aggregation strategies: For a genome sequence of length $l$, let $P$ denote the partition of the indices $\{1, 2, ..., l\}$ corresponding to the tokenization of the sequence, i.e., a split into disjoint subsets that consist of the consecutive indices of a single token. Then for the $j$-th token with corresponding set of indices $p_j\in P$, the nucleotide-level relevance $r_i^{(nucleo)}$ is aggregated by
    \begin{enumerate}
        \item[(a)] $r_j^{(token)}:=\sum_{i\in p_j}r_i^{(nucleo)}$ (sum aggregation) 
        \item[(b)] $r_j^{(token)}:=\frac{1}{|p_j|}\sum_{i\in p_j}r_i^{(nucleo)}$ (mean aggregation).
    \end{enumerate}
Conversely, for every nucleotide $i\in p_j$, the token relevance $r_j^{(token)}$ is disaggregated by
    \begin{enumerate}
        \item[(c)] $r_i^{(nucleo)}:=r_j^{(token)}$ (passed on)
        \item[(d)] $r_i^{(nucleo)}:=\frac{1}{|p_j|}r_j^{(token)}$ (equally distributed).
    \end{enumerate}

Strategies (a) and (d) preserve the total relevance sum, whereas (b) and (c) do not. While a conservation of relevance might be methodologically desirable, it can dilute relevance of important nucleotides. For example, when only a few nucleotides within a token are relevant, both averaging their relevance with less important neighbors or equally distributing their token relevance can diminish their actual importance.

\section{Experiments}

In our experiments, we evaluate the explanations of AttnLRP for two different downstream tasks of a widely used genomic benchmark dataset. Therefore, we use the implementation of AttnLRP in \texttt{lxt} \cite{achtibat2024attnlrp}. We compare AttnLRP explanations against LRP explanations of a CNN. This involves multiple metrics to evaluate the relevance attributions from different perspectives\footnote[1]{Code available at \url{https://gitlab.com/dacs-hpi/explain_dnabert2}}

\subsection{Evaluation Metrics}\label{sec_metrics}
To evaluate and compare the explanations of AttnLRP for the gLMs and of LRP for the CNNs, we apply multiple evaluation metrics. In particular, we consider how the explanations align, how complex they are, whether they actually capture features relevant for prediction, and compare what they identified with known biological annotations. They are described in more detail in the following paragraphs.

\paragraph{Similarity}
To evaluate how well two explanations align, we use the \textit{Continuous Jaccard Similarity} (CJ) introduced by \cite{shrikumar2018technical}. For two vectors $v,w\in\mathbb{R}^d$, it is defined as
\begin{align*}
        \mathrm{CJ}(v,w):= \frac{\sum_{i=1}^d \mathrm{sign}(v_i)\,\mathrm{sign}(w_i)\min\left(|v_i|,|w_i|\right)}{\sum_{i=1}^d \max\left(|v_i|, |w_i|\right)} \,.
\end{align*}

It ranges from $-1$ to 1 with $\mathrm{CJ}(v,w)=1$ if $v=w$ and $\mathrm{CJ}(v,w)=-1$ if $v=-w$. For relevance attributions, this means that scores with matching signs and magnitudes contribute positively to similarity, while mismatches reduce it.

\paragraph{Sparsity and Complexity}
To be interpretable, an explanation should avoid excessive complexity, in the sense that only a small subset of input features contributes meaningfully to the attribution. While it is known that post-hoc attributions can provide useful explanations for CNNs, this is not clear for gLMs due to their more complex model architecture. Therefore, we measure sparsity and complexity of explanations by Gini Index or entropy, respectively, following \cite{chalasani2020concise} and \cite{bhatt2020evaluating}. The Gini Index measures how equally the relevance is distributed among the nucleotides where for our purpose it is more desirable to have fewer input features with significant relevance. In contrast, higher entropy indicates a more diffuse (and thus more complex) explanation, while lower entropy corresponds to simpler, more concentrated explanations.
We use both metrics as implemented in the \texttt{quantus} toolkit \cite{hedstrom2023quantus}. 

\paragraph{Faithfulness}
An explanation is considered faithful if its assigned relevance scores actually correspond to the importance of the features for prediction. To assess this, we perturb the top $k$\% of tokens or nucleotides ranked by their attribution scores (most important first: MIF) \cite{samek2021explaining} and measure the difference in the prediction score. For DNABERT-2 we perturb on a token level and for the CNN on a nucleotide level, but in each perturbation round the same amount of nucelotides are perturbed. We consider different ways of perturbing the sequence: replacing the corresponding nucleotide position with the unknown token (UNK) or nucleotide (`N'), a random nucleotide, or the complement. Note that the random perturbation should be considered cautiously as it might reintroduce patterns into the sequence. As a sanity check, we also evaluate the effect of perturbing the least important $k$\% of features first (LIF) because, unlike MIF, this should lead to no or small changes in prediction. For both setups, we evaluate at $k \in \{1, 5, 10, 20, 50\}$ as proposed by \cite{deyoung-etal-2020-eraser}.

\paragraph{Localization (Database Comparison)}
Localization refers to whether the patterns identified by the explanation correspond to known annotations. To obtain motifs from the generated explanations, we use TF-MoDISco (Transcription Factor Motif Discovery from Importance Scores) \cite{shrikumar2018technical}. It takes relevance attributions as input, identifies regions of high importance, clusters and aggregates them, and outputs position weight matrices (PWMs). They represent a motif by providing frequencies or proportions for the four nucleotides at every position.
The PWMs are then compared against the motif database JASPAR \cite{sandelin2004jaspar} using TOMTOM \cite{gupta2007quantifying}, which aligns the discovered motifs to reference motifs and reports significant matches. 

\subsection{Dataset}
We use two benchmark datasets for genome sequence classification from the widely used collection curated by \cite{gresova2023genomic}. They combine experimentally validated regions (positive class) with randomly sampled genomic segments (negative class). The datasets already provide a train-test split. We use the train set for fine-tuning DNABERT-2 and conduct the experiments on the test set. To cover different sequence lengths and species, we select the following two datasets:

\paragraph{Nontata Promoters (Human)} Short sequences of length 251 base pairs (bp) representing non-TATA promoters vs. random coding regions. Promoters are regions in the sequence which are in proximity of genes initiating the reading (or transcription) of a gene. The dataset contains 36,131 samples with a positive-to-negative class ratio of approximately 1.2.
\paragraph{Drosophila Enhancers} Longer sequences (median length: 2,142 bp, SD: 285.5) representing validated enhancers vs. random genomic regions from the \textit{Dro\-so\-phila} genome. Enhancers are regions that regulate transcription. The dataset includes 6,914 samples and is balanced (class ratio 1.0).

\subsection{Models}\label{subsec_models}
For the \textit{nontata promoters} dataset, we adopt the baseline CNN provided with the benchmark dataset \cite{gresova2023genomic} achieves a test accuracy of 84.6\%.

For the \textit{drosophila enhancers} dataset, the baseline CNN reaches only 58.6\% test accuracy. To enhance performance, we implemented the model DeepSTARR \cite{de2022deepstarr}, a CNN originally designed for \textit{Drosophila} data. We extended its input length from 1,001 to 3,061 bp (matching the longest sequence in the dataset). After being trained on the \textit{drosophila enhancers} dataset, the model achieves 77\% test accuracy.

DNABERT-2 is fine-tuned for both classification tasks using the Hugging Face Trainer \cite{wolf2020transformers}. Input sequences are tokenized with the pretrained DNABERT-2 BPE tokenizer \cite{zhou2023dnabert}. The final classifiers achieve test accuracies of 94\% on \textit{nontata promoters} and 78\% on \textit{drosophila enhancers}.

\section{Results}
\begin{figure}
    \centering
    \includegraphics[width=1\linewidth]{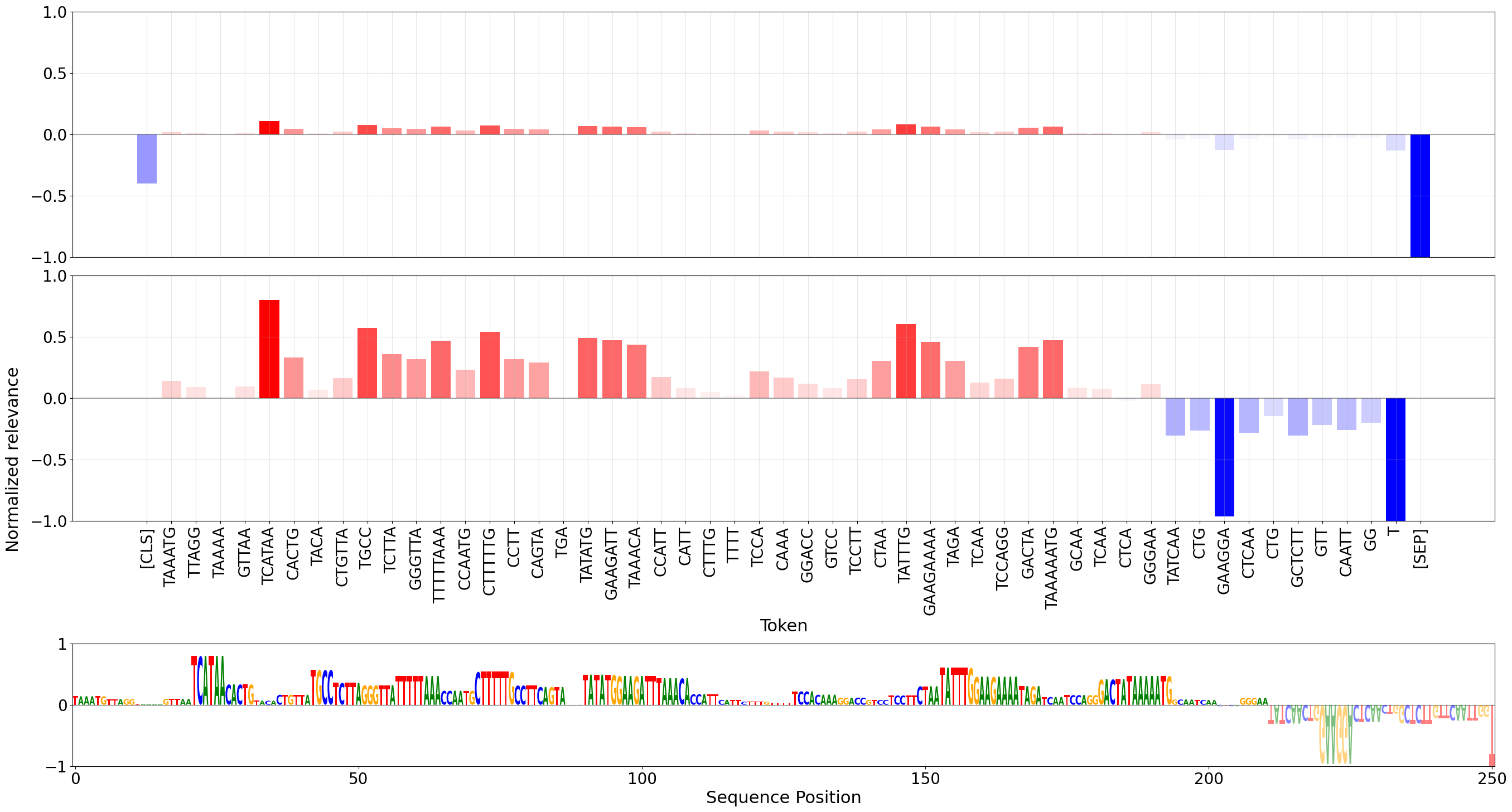}
    \caption{Relevance attribution of DNABERT-2 by AttnLRP for a sample of the \textit{nontata promoters} dataset. The first row displays the attribution including CLS and SEP who receive by far the largest relevance in absolute terms. In the second row, CLS and SEP were excluded and the relevance attribution was re-normalized. This makes the patterns in between visible. The third row shows the corresponding logo plot (letters show nucleotide and height of the letters relevance scores) with disaggregation strategy (c) (passed on).}
    \label{fig:attnmaps}
\end{figure}

We start with some general observation before we discuss the results regarding the evaluation metrics in the later paragraphs. 
In a first qualitative inspection of the adapted AttnLRP explanations of DNA\-BERT-2, we see relevance attributions with visible non-random patterns (see Figure \ref{fig:attnmaps}), i.e., consecutive tokens with high or low relevance (in absolute terms) indicating important or non-important regions in the genome sequence. Hence, our adaptation of AttnLRP seems to provide explanations of DNABERT-2 that are clear enough to draw conclusions from them despite the model's increased complexity compared to CNNs. Furthermore, we notice some peculiarities concerning the CLS and SEP tokens.
When applying AttnLRP to a BERT-based model, we consistently observe that a substantial portion of the relevance is concentrated at the special tokens CLS and SEP. The remaining relevance is distributed among the other tokens. Hence, as the input sequence length increases, the relevance assigned to non-special tokens decreases. 
This observation is in line with relevance attributions of a BERT model for natural language provided in the documentation of the \texttt{lxt} package \cite{achtibat2024attnlrp}. They use a shorter example sentence where the effect is weaker. If we increase the sentence length, we observe the same effect as in DNABERT-2.
However, we are interested in the relevance of nucleotides in the sequence, i.e. represented by the tokens between CLS and SEP. Therefore, we exclude CLS and SEP from the attribution analysis. Following their exclusion, the relevance scores of the remaining tokens are re-normalized to span the interval $[-1, 1]$, thereby preserving the relative distribution of relevance while ensuring comparability across sequences of different lengths.

\begin{figure}[h]
    \centering
    \includegraphics[width=\linewidth]{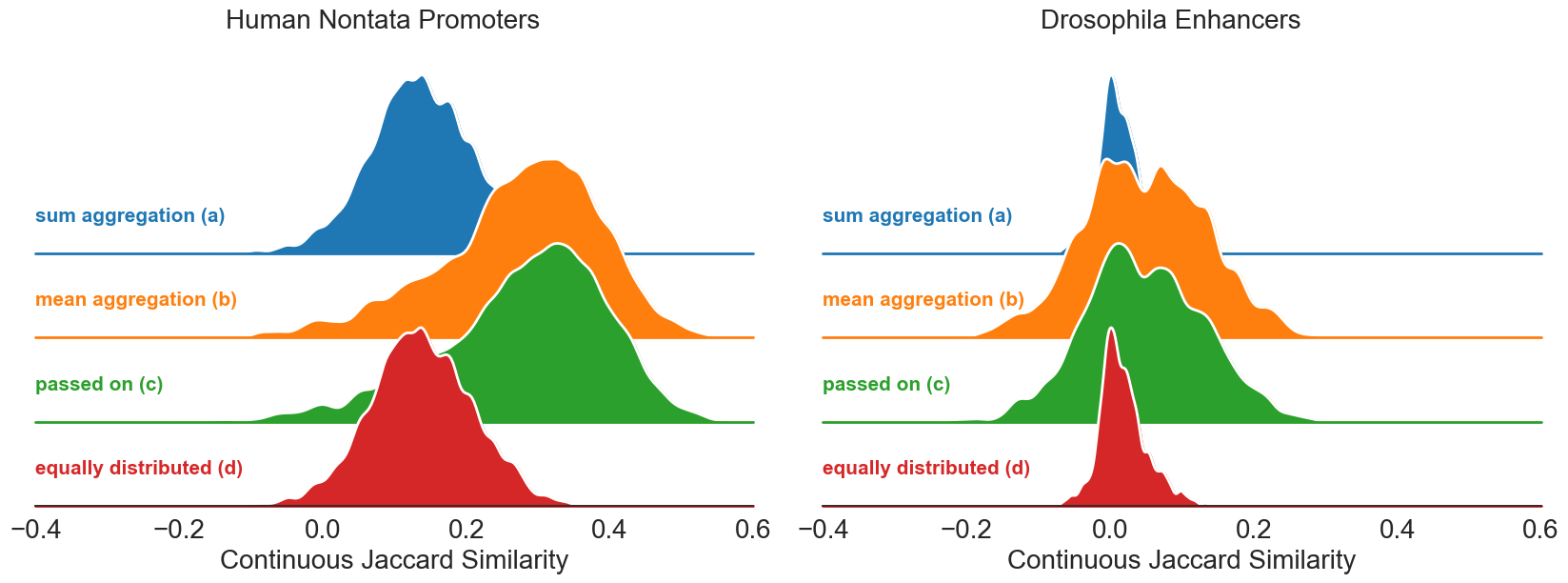}
    \label{fig:img1}
    \caption{Distribution of the continuous Jaccard similarity between explanations of DNABERT-2 and the baseline CNN for positive-class predictions, shown for the Human Nontata Promoters dataset (left) and the Drosophila Enhancers dataset (right). Each ridge corresponds to a different tokenization (dis-)aggregation strategy. For Human Nontata Promoters, all strategies yield distributions centered around positive similarity values, indicating moderate agreement between model explanations. In contrast, the Drosophila Enhancers dataset shows distributions closer to zero, suggesting lower correspondence between DNABERT-2 and CNN attributions.}
    \label{fig:combined}
\end{figure}

\paragraph{Similarity}
Figure \ref{fig:combined} illustrates the distribution of the continuous Jaccard similarity (CJ) between the explanations of DNABERT-2 and the baseline CNN for predicting the positive class for the (dis-)aggregation strategies.
For both datasets, we observe higher CJ values for strategies (b) and (c) compared to (a) and (d). This indicates that although (a) and (d) are relevance-preserving they might dilute relevance for important nucleotides and hence, align worse than the non-sum-preserving transformations (b) and (c). For the \textit{nontata promoters} dataset, we observe a mostly positive CJ, which indicates that both models have a certain overlap in the motifs they use for prediction. For the \textit{drosophila enhancers} dataset, we see a much weaker alignment, i.e., lower positive and more negative values for CJ. This is partially expected as the dataset consists of longer sequences and hence, the explanations should have a lower signal-to-noise ratio. Overall, for some samples, the explanations of DNABERT-2 and the CNN seem to align to a certain level. On the other hand, for samples with a CJ close to zero, the models seem to rely on different features, although both predict the positive class. For the negative class, we see lower alignment (data omitted here). This is expected as these samples were likely classified based on the absence of motifs and hence, there might be fewer patterns the model can agree on.  

\paragraph{Sparsity and Complexity}
\label{subsec_spars}
We assessed the explanatory properties using sparsity and complexity metrics across both native and cross-granularities (see Table \ref{tab:sparsity_complexity_summary}). Complexity is inherently dependent on dimensionality; consequently, token-level explanations consistently exhibit lower complexity than nucleotide-level explanations (e.g., entropy of 3.63 vs. 5.22 or 3.49 vs. 5.06 for \textit{nontata promoters}). Sparsity, which measures the inequality of attribution magnitudes, proved to be more robust to granularity changes but sensitive to sequence length. For the \textit{nontata promoters} dataset (short sequences), LRP on the CNN produced sparser explanations than AttnLRP on DNABERT-2 (Gini Index of 0.51 vs. 0.42). Conversely, for the \textit{drosophila enhancers} dataset (long sequences), AttnLRP yielded higher sparsity (0.64) compared to the CNN (0.50).

Overall, we observe that AttnLRP applied to DNABERT-2 provides explanations of similar sparsity and complexity as LRP explanations of CNNs. 
\renewcommand{\arraystretch}{1.1}
\setlength{\tabcolsep}{3pt}
\begin{table}
\centering
\caption{Means (and SDs) for sparsity and complexity by model and granularity across both datasets.Bold values mark best result for dataset and metric.}
\label{tab:sparsity_complexity_summary}
\begin{tabular}{lrrrr}
\toprule
\multicolumn{1}{c}{\multirow{2}{*}{Model}}& \multicolumn{2}{c}{Nontata Promoters} & \multicolumn{2}{c}{Drosophila Enhancers} \\
\cmidrule(lr){2-5} & \multicolumn{1}{c}{Gini Index} & \multicolumn{1}{c}{Entropy} & \multicolumn{1}{c}{Gini Index} & \multicolumn{1}{c}{Entropy} \\
\midrule
DNABERT-2 token-level      & 0.42 (0.07) & 3.63 (0.13) & \textbf{0.64} (0.10) & \textbf{5.02} (0.56)\\
DNABERT-2 nucleotide-level & 0.41 (0.07) & 5.22 (0.12) & \textbf{0.64} (0.10) & 6.63 (0.55)\\
CNN token-level          & \textbf{0.51} (0.07) & \textbf{3.49} (0.16) & 0.50 (0.05) & 5.57 (0.27)\\
CNN nucleotide-level     & \textbf{0.51} (0.07) & 5.06 (0.14) & 0.50 (0.05) & 7.17 (0.27) \\
\bottomrule
\end{tabular}
\end{table}

\begin{figure}
\includegraphics[width=\textwidth]{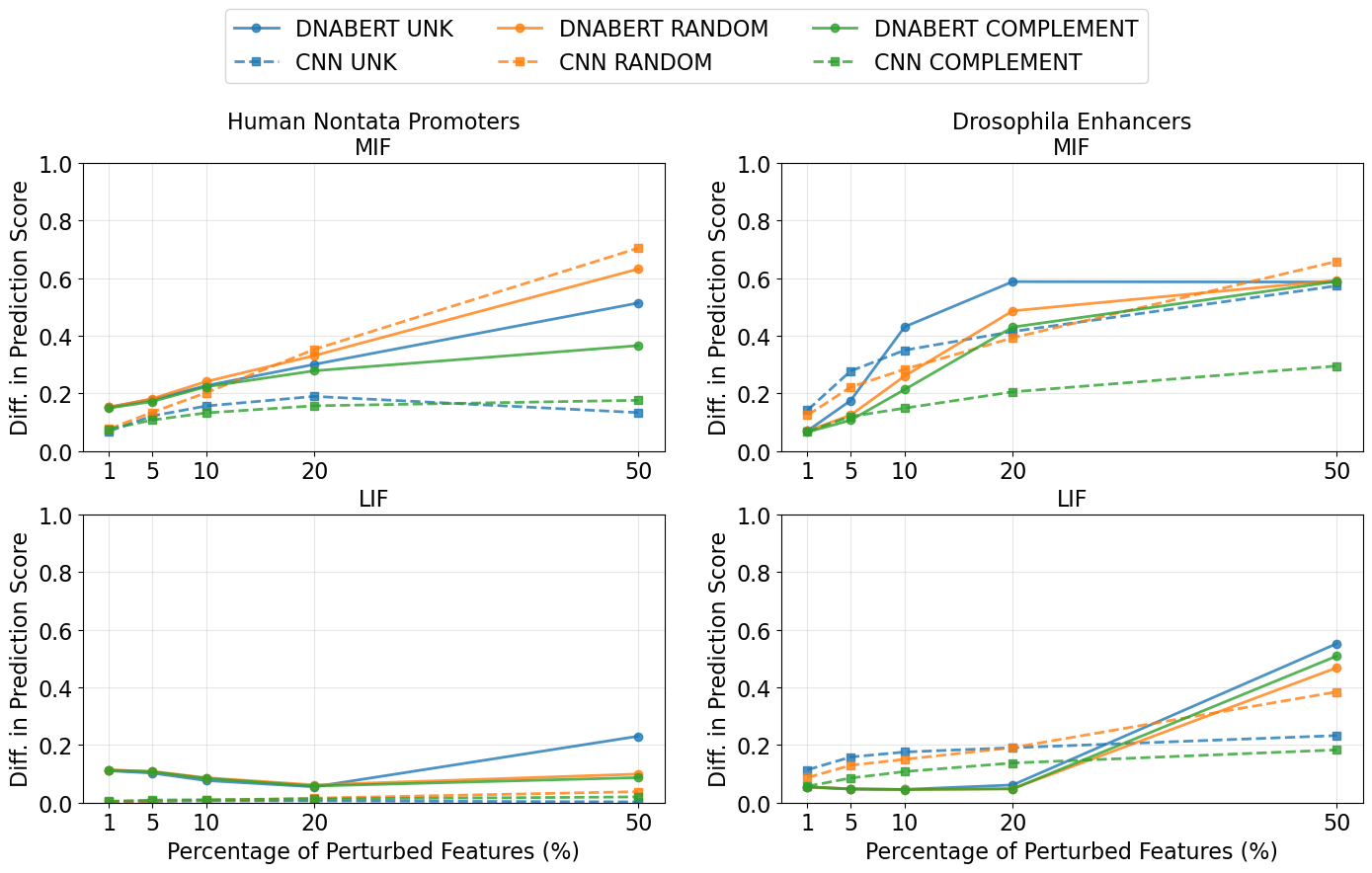}
\caption{Results for faithfulness: difference in prediction scores for the positive class after perturbing input features according to MIF (upper) and LIF (lower) with unknown, random and complement nucleotides for nontata promoters (left) and drosophila enhancers dataset (right).} \label{fig_results_faithfulness}
\end{figure}

\paragraph{Faithfulness}
\label{subsec_comp}
For faithfulness, we consider the change in the model's prediction score after perturbing the input features according to the relevance attribution. A larger difference in the prediction score indicates that the perturbed features were indeed relevant for prediction. Figure \ref{fig_results_faithfulness} illustrates the results for the positive class. For perturbing features with unknown or complement nucleotides according to MIF, we observe that the DNABERT-2 explanations generally yield higher values (see blue and green lines in upper row) than the baseline CNN explanations. Hence, AttnLRP is capable of identifying features that are important for the predictions of DNABERT-2 similar as LRP does for the CNN.
For the random perturbation, we observe slightly higher values for the baseline when perturbing 50\% of the features. However, as mentioned in section \ref{sec_metrics}, random perturbation can reintroduce patterns and hence, distort the scores. Furthermore, for the \textit{drosophila enhancers} dataset, the CNN seems to be more sensitive when perturbing only a few features indicated by the higher scores for the baseline. However, this vanishes when perturbing more nucleotides. When perturbing up to 20\% of the features according to LIF (lower row), the AttnLRP explanations lead to similar small changes in the prediction score as the CNN explanations. This indicates that the nucleotides with low relevance assigned by AttnLRP also indeed have little importance for the prediction of DNABERT-2.
We present here only the results for the positive class. For both datasets, the task is to find patterns in the sequence. Hence, while explanations of the negative class can still provide useful insights, e.g., the absence of a motif at a position in the sequence where the model would expect one for the positive class, perturbing them in general does not lead to a class flip and hence, not to a drop in the prediction score. In fact, we observe these flat curves when considering the negative class, but omit them here.
Overall, we can conclude that AttnLRP provides similarly faithful explanations of DNABERT-2 as LRP does for a CNN.

\paragraph{Localization (Database comparison)}
\label{subsec_database}
To validate the accordance of the explanations with known biological patterns, we compare them to the JASPAR motif database. For the database comparison we need nucleotide level relevance scores. As we observe similar output patterns of the TF-MoDISco algorithm and motif matches for strategies (c) and (d), we report only the results of (c) in the following. Due to the sparsity and variable length of the \textit{drosophila} sequences, standard TF-MoDISco parameters required adjustment (window tightening and cluster resizing) to avoid unstable cosine normalizations.
Every pattern extracted via TF-MoDISco from the DNABERT-2 explanations matched a known motif, indicating that the model relies on biologically plausible regions. Note that these regions can also be spurious motifs or context patterns that the model leverages for prediction.
\begin{figure}
    \centering
    \includegraphics[width=\linewidth, height=7cm, keepaspectratio]{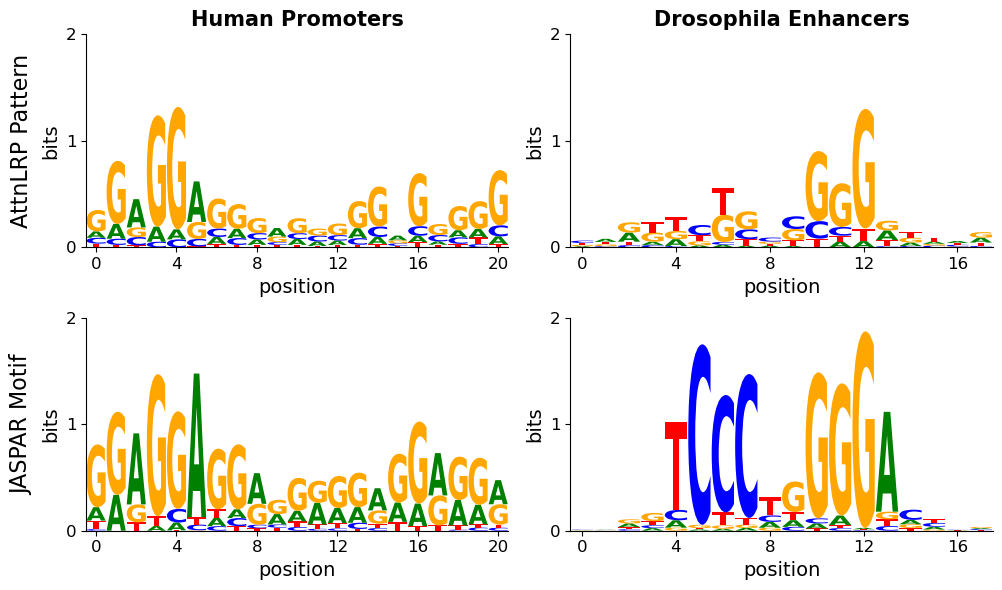}

    \caption{Database comparison: best match of pattern of positive class in the AttnLRP patterns with database target for nontata promoters dataset and drosophila enhancers dataset. Patterns generated by TF-MoDISco. The results show a higher overlap of the pattern and the target motif for the human nontata promoters dataset. Letter height represents information content in bits, quantifying how much a position deviates from random expectation; 2 bits indicates complete conservation.}
    \label{fig_results_comparison}
\end{figure}
For the \textit{nontata promoters} dataset, the DNABERT-2 explanations highlight GC-boxes---areas with high amounts of G and C bases (e.g., motif MA0528.1, see Figure \ref{fig_results_comparison}). This aligns with the higher GC-content observed in the positive class (62\%) versus the negative class (48\%), suggesting the models discriminate based on GC-rich regions. For the baseline CNN, we also see a focus on GC-boxes (omitted here). While the CNN extracted a higher quantity of patterns, the patterns derived from DNABERT-2 were often more concentrated on fewer nucleotides. 

For the \textit{drosophila enhancers} dataset, explanations of both models converged to the same top motif (samples with high similarity in relevance maps): MA1841.1 (related to enhancer repression). This convergence suggests that the Transformer-based architecture and the CNN capture the same underlying biological signals despite operating at different granularities. The identification of known repressive motifs in the negative class further confirms that the models are leveraging biologically plausible features rather than confounding artifacts to make predictions.

\section{Conclusion}
In this work, we adapted AttnLRP to DNABERT-2. For this, we transferred the gradient division to the ALiBi and GLU components in DNABERT-2. This allowed us to compute explanations of a Transformer-based gLM. Further, we proposed four (dis-)aggregation strategies to transform explanations on token level to nucleotide level and vice versa. These strategies allowed us to compare explanations of token-based Transformer models with those of nucleotide-based CNNs. Our evaluation showed that AttnLRP provides similar reliable explanations of DNABERT-2 as LRP does for CNNs. This is an important result for the community because it has not been clear whether Transformer-based gLMs provide reliable explanations despite their more complex architecture. Furthermore, we could verify that the highlighted patterns correspond to known biological annotations. Although this only confirms existing knowledge, this shows that explanations of Transformer-based gLMs can potentially discover new biological insights, and hence be used for hypothesis generation. Since we focused on DNABERT-2 only, it remains an open question whether these results transfer to other Transformer-based models trained on genome sequences. Additionally, since we only considered two datasets, it should be further investigated how the type of data, e.g. sequence length, species, prediction task, influences the explanations. Hence, in future work, we aim to extend the analysis to other models and datasets and consider explanation methods beyond AttnLRP to further contribute to explainability of deep learning in genomics.

\subsubsection*{Acknowledgements}
We gratefully acknowledge funding by grant RE 3474/8-1 (to BYR), project P5 in the Research Unit KI-FOR 5363 (grant 459422098) of the German Research Foundation (DFG).

We thank Marta Lemanczyk and Simon Witzke for reviewing the manu\-script and providing helpful comments.

\subsubsection*{Disclosure of Interests}
The authors have no competing interests to declare that are
relevant to the content of this article.

\subsubsection*{Use of Generative AI}
The authors used ChatGPT (GPT-4 and GPT-5 by OpenAI) for language editing of portions of the manuscript and for assistance with Python code development. All AI-generated content was carefully reviewed and edited by the authors, who take full responsibility for the accuracy, originality, and integrity of the manuscript. Generative AI was not used for data analysis, interpretation of results, or generation of scientific conclusions.

\bibliographystyle{splncs04}
\bibliography{references}

\end{document}